\documentclass[pdflatex,sn-mathphys-num]{sn-jnl}

\usepackage{graphicx}%
\usepackage{multirow}%
\usepackage{amsmath,amssymb,amsfonts}%
\usepackage{amsthm}%
\usepackage{mathrsfs}%
\usepackage[title]{appendix}%
\usepackage{xcolor}%
\usepackage{textcomp}%
\usepackage{manyfoot}%
\usepackage{booktabs}%
\usepackage{algorithm}%
\usepackage{algorithmicx}%
\usepackage{algpseudocode}%
\usepackage{listings}%

\usepackage{mathtools}

\theoremstyle{thmstyleone}%
\theoremstyle{thmstyletwo}%

\theoremstyle{thmstylethree}%

\begin{document}

\title[Moving Sofa by AI]{Deep Learning Evidence for Global Optimality \\of Gerver's Sofa}

\author*[1]{\fnm{Kuangdai} \sur{Leng}}\email{kuangdai.leng@stfc.ac.uk}

\author[1]{\fnm{Jia} \sur{Bi}}\email{jia.bi@stfc.ac.uk}

\author[1]{\fnm{Jaehoon} \sur{Cha}}\email{jaehoon.cha@stfc.ac.uk}

\author[1]{\fnm{Samuel} \sur{Pinilla}}\email{samuel.pinilla@stfc.ac.uk}

\author[1]{\fnm{Jeyan} \sur{Thiyagalingam}}\email{t.jeyan@stfc.ac.uk}

\affil[1]{\orgdiv{Scientific Computing Department}, \orgname{Science and Technology Facilities Council}, \orgaddress{\street{Rutherford Appleton Laboratory}, \city{Didcot}, \postcode{OX11 0QX}, \country{UK}}}

\abstract{The Moving Sofa Problem, formally proposed by Leo Moser in 1966, seeks to determine the largest area of a two-dimensional shape that can navigate through an $L$-shaped corridor with unit width. The current best lower bound is about 2.2195, achieved by Joseph Gerver in 1992, though its global optimality remains unproven. In this paper, we investigate this problem by leveraging the universal approximation strength and computational efficiency of neural networks. We report two approaches, both supporting Gerver's conjecture that his shape is the unique global maximum. Our first approach is continuous function learning. We drop Gerver's assumptions that i) the rotation of the corridor is monotonic and symmetric and, ii) the trajectory of its corner as a function of rotation is continuously differentiable. We parameterize rotation and trajectory by independent piecewise linear neural networks (with input being some pseudo time), allowing for rich movements such as backward rotation and pure translation. We then compute the sofa area as a differentiable function of rotation and trajectory using our ``waterfall'' algorithm. Our final loss function includes differential terms and initial conditions, leveraging the principles of physics-informed machine learning. Under such settings, extensive training starting from diverse function initialization and hyperparameters is conducted, unexceptionally showing rapid convergence to Gerver's solution. Our second approach is via discrete optimization of the Kallus-Romik upper bound, which converges to the maximum sofa area from above as the number of rotation angles increases. We uplift this number to 10000 to reveal its asymptotic behavior. It turns out that the upper bound yielded by our models does converge to Gerver's area (within an error of 0.01\% when the number of angles reaches 2100). We also improve their five-angle upper bound from 2.37 to 2.3337.}




\maketitle

\section{Introduction}
\begin{figure}[h]
 \centering
 \includegraphics[width=\textwidth]{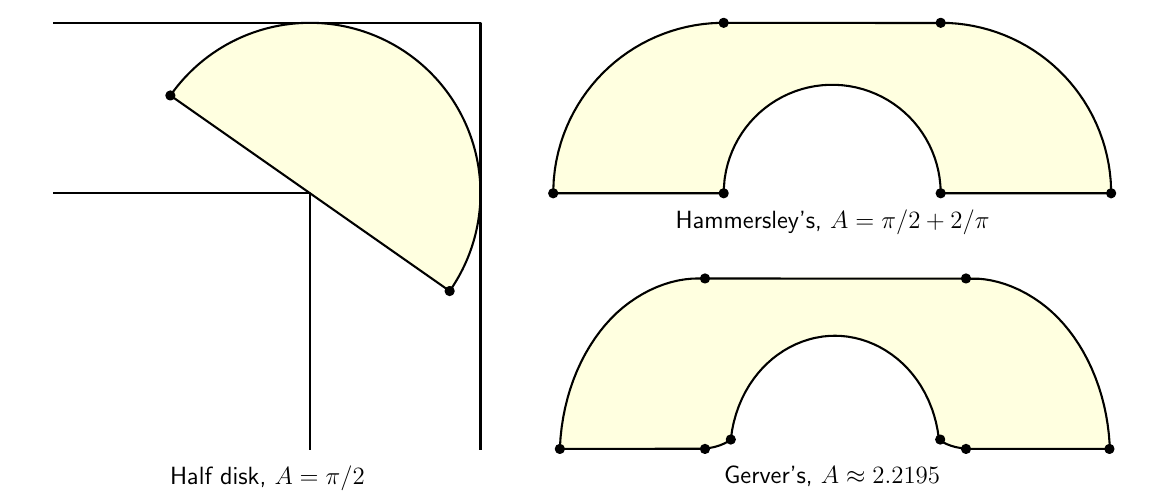}
 \caption{The moving sofa problem proposed by Leo Moser~\cite{moser1966problem} in 1966 and some lower bounds. The best-known two are found respectively by John Hammersley~\cite{hammersley1968enfeeblement} and Joseph Gerver~\cite{gerver1992moving}. The point markers separate the sections formed by different contact mechanisms (i.e., with either the inner corner, the walls, or the wall envelopes of the corridor).}
 \label{fig:intro}
\end{figure}

\begin{figure}[b]
	\centering
	\includegraphics[width=\textwidth]{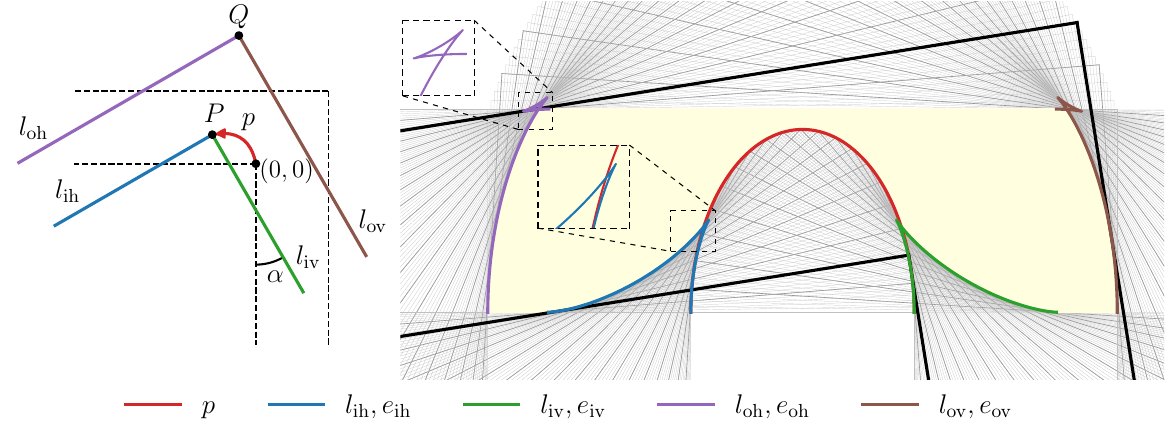}
	\caption{Geometry of the moving sofa problem. The movement of the corridor is described by the trajectory of its inner corner $P$, as denoted by $p$, and the rotation angle $\alpha$. The four walls form the four families of lines: $l_\text{ih}$, $l_\text{iv}$, $l_\text{oh}$ and $l_\text{ov}$, with the subscripts showing their initial positions (i for inner, o for outer, h for horizontal and v for vertical). Their envelopes are respectively $e_\text{ih}$, $e_\text{iv}$, $e_\text{oh}$ and $e_\text{ov}$. In this example, $p$ is a semi-ellipse with its major and minor lengths being respectively 1.8 and 1.1, leading to some complexities in the envelopes, as highlighted by the magnified windows.  }
	\label{fig:geo}
\end{figure}

Among the unsolved problems in geometry, the moving sofa problem stands out for its apparent simplicity. Formally proposed by Leo Moser~\cite{moser1966problem}, it asks for the two-dimensional shape of the largest area that can be maneuvered through an $L$-shaped corridor with unit width (see Figure~\ref{fig:intro}). Beyond trivial shapes such as square and half disk, the best-known lower bounds are those found by John Hammersley~\cite{hammersley1968enfeeblement} and Joseph Gerver~\cite{gerver1992moving}, as shown in Figure~\ref{fig:intro}. Hammersley's sofa, as he presented in a problem set for school and college students, has an area of $\pi/2+2/\pi\approx2.2074$. It is the largest among the family whose touch point with the inner corner of the corridor forms a semi-circle\footnote{It is then natural to ask if one can generalize semi-circle to semi-ellipse, which has sparked valuable efforts from enthusiasts~\cite{geogbra,1787466}; as illustrated in our Figure~\ref{fig:geo}, however, such generalization is not straightforward.}; he also showed an upper bound of $2\sqrt2\approx2.8284$. With an area slightly larger than $2.2195$, Gerver's sofa is the largest known to date, composed of 18 curve sections (as partitioned by their analytical expressions). In a fairly implicit way, Gerver proved that his solution is a local maximum and conjectured that it is the only global maximum.

Dan Romik~\cite{romik2018differential} advanced Gerver's analysis by introducing a set of six differential equations along with their general solutions. These solutions, under prescribed contact mechanisms through five phases of rotation, land on Gerver's sofa in a natural and explicit way. Using the same method, Romik also discovered a lower bound of the ``ambidextrous sofa'' that can pass through a double-cornered corridor. A numerical study~\cite{batsch2022numerical} was carried out based on Romik's formulation. Yoav Kallus and Dan Romik~\cite{kallus2018improved} illuminated a new perspective of the problem. They proved an upper bound of the maximum area through discrete optimization of the rotation center at a finite sequence of rotation angles. Under a seven-angle setup, they discovered a new upper bound of the area at 2.37, along with the least-required rotation for maximum area production by approximately $81.2^\circ$. We will delve into their findings in greater detail in Section~\ref{sec:KR}.

The past decade has witnessed groundbreaking advancements in deep learning, revolutionizing numerous facets of human endeavor, including mathematical research, an area now often referred to as \textit{AI4Math}. In the realm of constructing examples or counterexamples (e.g., finding a new moving sofa is one of such), some AI-discovered results have surpassed human efforts, such as faster algorithms for matrix production~\cite{fawzi2022discovering}, larger cap sets for the no-three-in-line problem~\cite{romera2024mathematical}, and many complex knots with specific topological properties~\cite{gukov2023searching}. Here we seek to probe the moving sofa problem through the lens of deep learning. One of the core techniques we will employ is physics-informed neural networks (PINNs), dating back to 90's~\cite{dissanayake1994neural,lagaris1998artificial} and now thriving in the contemporary deep learning ecosystem~\cite{raissi2019physics, karniadakis2021physics}. PINNs are aimed mainly at solving and inverting differential equations, featuring the idea of incorporating the target equations, along with their initial and boundary conditions, into the network architecture or loss function. In general, PINNs can enhance accuracy and generalization of the resultant models while reducing the amount of training data required. We will adhere to the principles of PINNs to address the differential terms involved in the geometry and the initial conditions of sofa movement. By doing so, we aim to achieve more stable and efficient training while imposing minimum assumptions on the movement.

We approach the moving sofa problem from an optimization viewpoint, employing neural networks (NNs) to parameterize a function space for identifying optimal sofa movements or upper bounds. The idea harnesses the universal approximation capability of NNs and the enhanced, albeit still evolving, proficiency of modern optimizers to avoid local minima. While preliminary attempts from the community exist (e.g.,~\cite{Luna2022}), they often fail to properly consider the geometry and lack comprehensive insights into optimality. A significant challenge we have surmounted is the computation of the sofa area, which engages with complex (or even chaotic) geometry given an arbitrary movement. To this end, we propose the ``waterfall'' algorithm, which can compute the area not only with high accuracy and robustness but also in a differential manner as required for backpropagation.

Two independent approaches will be described in the remainder of this paper. In Section~\ref{sec:area}, we directly optimize the sofa area as a function of corridor movement, parameterized by PINNs. To minimize inductive bias, special attention is paid to the generality of the function space and the diversity of function initialization. In Section~\ref{sec:KR}, we shift our objective function from the sofa area to the Kallus-Romik upper bound~\cite{kallus2018improved}. We first revisit their five-angle scenario, obtaining a tighter upper bound at 2.3337, and then explore a large number of angles to show the asymptotic behavior of their upper bound. Initially motivated to detect a shape larger than that of Gerver's, both our two approaches end up being supportive of Gerver's conjecture that his 18-section sofa is the unique global maximum.

\section{Area optimization}
\label{sec:area}

\subsection{Geometry}
\label{sec:geo}
The moving sofa problem can be understood as determining the movement of the $L$-shaped corridor in $\mathbb{R}^2$ to maximize the area not swept by its four walls. We describe its movement using the trajectory of its inner corner, denoted by $p$, and the rotation angle $\alpha$, as illustrated in Figure~\ref{fig:geo}. This representation allows to describe the corridor's position and orientation at any moment. An interactive visualization, assuming an elliptical trajectory for $p$, has been contributed by the community~\cite{geogbra}. Previous studies by continuous optimization have considered $\left(x_p, y_p\right)$ as functions of $\alpha$~\cite{gerver1992moving, romik2018differential, batsch2022numerical}, thereby restricting the movement to a monotonic rotation. To overcome this limitation, we introduce a pseudo-time $t\in\left[0,1\right]$, and describe the movement by
\begin{equation}
 \begin{cases}
 \begin{aligned}
 x_p&=x_p\left(t\right),\\
 y_p&=y_p\left(t\right),\\
 \alpha&=\alpha\left(t\right);
 \end{aligned}
 \end{cases}
 \quad \text{satisfying} \quad 
 \begin{cases}
 \begin{aligned}
 x_p\left(0\right)&=0,\\
 y_p\left(0\right)&=0,\\
 \alpha\left(0\right)&=0.
 \end{aligned}
 \end{cases}
 \label{eq:par}
\end{equation}
The above initial conditions at $t=0$ are trivial, dealt with the translation and rotation invariance of $\mathbb{R}^2$. A non-trivial condition is 
\begin{equation}
 \alpha\left(1\right)>\arcsin\frac{84}{85}\approx81.2^\circ,
 \label{eq:81}
\end{equation}
that is, any moving sofa shape of largest area must undergo rotation by an angle of at least $81.2^\circ$~\cite{kallus2018improved}. We will later use this condition in our loss function. The parameterization in Eq.~\eqref{eq:par} only requires that $x_p\left(t\right)$, $y_p\left(t\right)$ and $\alpha\left(t\right)$ are of class $C^0$, allowing for independent translation and rotation that are non-monotonic, non-differentiable and non-symmetric.

The movement of the corridor will determine the following four families of lines, as named by their initial position at $t=0$ (see Figure~\ref{fig:geo}):
\begin{equation}
 \begin{aligned}
 \text{Inner horizontal, } l_\text{ih}: &\quad
 \left(x-x_p\right)\sin\alpha-\left(y-y_p\right)\cos\alpha=0;\\
 \text{Inner vertical, } l_\text{iv}: &\quad
 \left(x-x_p\right)\cos\alpha+\left(y-y_p\right)\sin\alpha=0;\\
 \text{Outer horizontal, } l_\text{oh}: &\quad
 \left(x-x_p\right)\sin\alpha-\left(y-y_p\right)\cos\alpha+1=0;\\
 \text{Outer vertical, } l_\text{ov}: &\quad
 \left(x-x_p\right)\cos\alpha+\left(y-y_p\right)\sin\alpha-1=0.
 \end{aligned}
\end{equation}
Based on simple calculus, their envelopes can be shown as
\begin{equation}
\begin{aligned}
 \text{Inner horizontal, } e_\text{ih}: &\quad
 \begin{cases}
 \begin{aligned}
 x_\text{ih} &= x_p+\cos\alpha \left(x_p'\sin \alpha-y_p'\cos\alpha\right)\left(\alpha'\right)^{-1},\\
 y_\text{ih} &= y_p+\sin\alpha \left(x_p'\sin \alpha-y_p'\cos\alpha\right)\left(\alpha'\right)^{-1};
 \end{aligned}
 \end{cases}
 \\
 \text{Inner vertical, } e_\text{iv}: &\quad
 \begin{cases}
 \begin{aligned}
 x_\text{iv} &= x_p-\sin\alpha \left(x_p'\cos \alpha+y_p'\sin\alpha\right)\left(\alpha'\right)^{-1},\\
 y_\text{iv} &= y_p+\cos\alpha \left(x_p'\cos \alpha+y_p'\sin\alpha\right)\left(\alpha'\right)^{-1};
 \end{aligned}
 \end{cases}\\
 \text{Outer horizontal, } e_\text{oh}: &\quad
 \begin{cases}
 \begin{aligned}
 x_\text{oh} &= x_\text{ih} - \sin\alpha,\\
 y_\text{oh} &= y_\text{ih} + \cos\alpha;
 \end{aligned}
 \end{cases}\\
 \text{Outer vertical, } e_\text{ov}: &\quad
 \begin{cases}
 \begin{aligned}
 x_\text{ov} &= x_\text{iv} + \cos\alpha,\\
 y_\text{ov} &= y_\text{iv} + \sin\alpha,
 \end{aligned}
 \end{cases}
\end{aligned}
\label{eq:env}
\end{equation}
where the primes denote derivatives with respect to $t$. We assume that $x_p\left(t\right)$, $y_p\left(t\right)$ and $\alpha\left(t\right)$ are piecewise differentiable (piecewise $C^1$) so that any non-differentiable points can be excluded from the aforementioned envelopes. This assumption is justified, as it is unlikely that the largest sofa could contain some fractal sections. If it does indeed, such sections are inherently undetectable by methods based on function parameterization, including deep learning approaches.

\begin{figure}
 \centering
 \includegraphics[width=\textwidth]{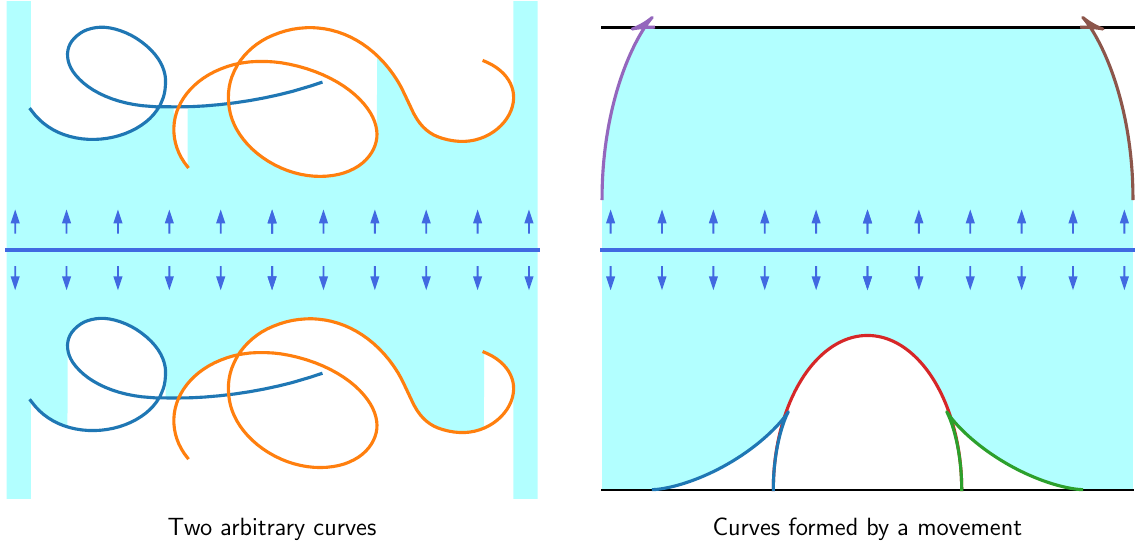}
 \caption{The waterfall algorithm for area calculation. Left: The waterfall is tested on two hand-drawn curves from both below and above. Right: The waterfall is applied to the curves formed by the corridor movement in Figure~\ref{fig:geo}. Water sources are placed along the horizontal line in the middle, with the arrows indicating the falling direction.}
 \label{fig:water}
\end{figure}

The resultant area is fringed from below by $p$, $e_\text{ih}$, $e_\text{iv}$ and $y=0$, whichever on top, and from above by $e_\text{oh}$, $e_\text{ov}$ and $y=1$, whichever on bottom. However, it cannot be formulated as a simple integral because these curves can be self-intersecting and intersect at multiple points, as indicated by the zoomed-in windows in Figure~\ref{fig:geo}. Notice that Figure~\ref{fig:geo} only shows a simple case where the trajectory is elliptical; the envelopes can become ``chaotic'' for an arbitrary movement. Such complexity also explains why the moving sofa problem remains unsolved. To compute the area robustly for any movements, we develop the ``waterfall'' algorithm, inspired by the rainfalling watershed algorithm for image segmentation~\cite{de2000implementation}. Intuitively, a dense series of water sources aligned horizontally are placed high above, generating a waterfall that touches the intersecting curves to form the lower limit for area quadrature. Our implementation ensures the differentiability of the resultant area with respect to the curve positions (for backpropagation) and parallelizes the water sources (by tensorization) for fast computation. Some examples are given in Figure~\ref{fig:water}.

\subsection{Physics-informed function learning}
\label{sec:learning}
Distinct from common deep learning tasks for data fitting, our application is not concerned with the accuracy or generalization of one best model. Instead, we aim to support the global optimality of Gerver's solution by demonstrating that many models from our over-parameterized function space converge to it. Therefore, we want our function space (i.e., hypothesis class) to be as general as possible. We parameterize $x_p\left(t\right)$, $y_p\left(t\right)$ and $\alpha\left(t\right)$ by three independent fully-connected networks (FCNs) using rectified linear unit (ReLU) as the activation function. 
Here FCNs are chosen for universal approximation~\cite{park1991universal,lu2021learning}, and ReLU for piecewise linear parameterization~\cite{tao2022piecewise,leng2022compatibility}. 
As explained after Eq.~\eqref{eq:env}, piecewise $C^{1}$ is the most general class of functions that can be explored using non-fractal methods; however, without any established piecewise $C^{1}$ architecture in deep learning, piecewise linearity, as facilitated by the ReLU family, strikes a balance between generality and practicality. 
In contrast, using $C^\infty$ activation functions such as $\tanh$ and $\mathrm{softplus}$ will narrow down the function space to $C^\infty$.

\begin{figure}
 \centering
 \includegraphics[width=.95\textwidth]{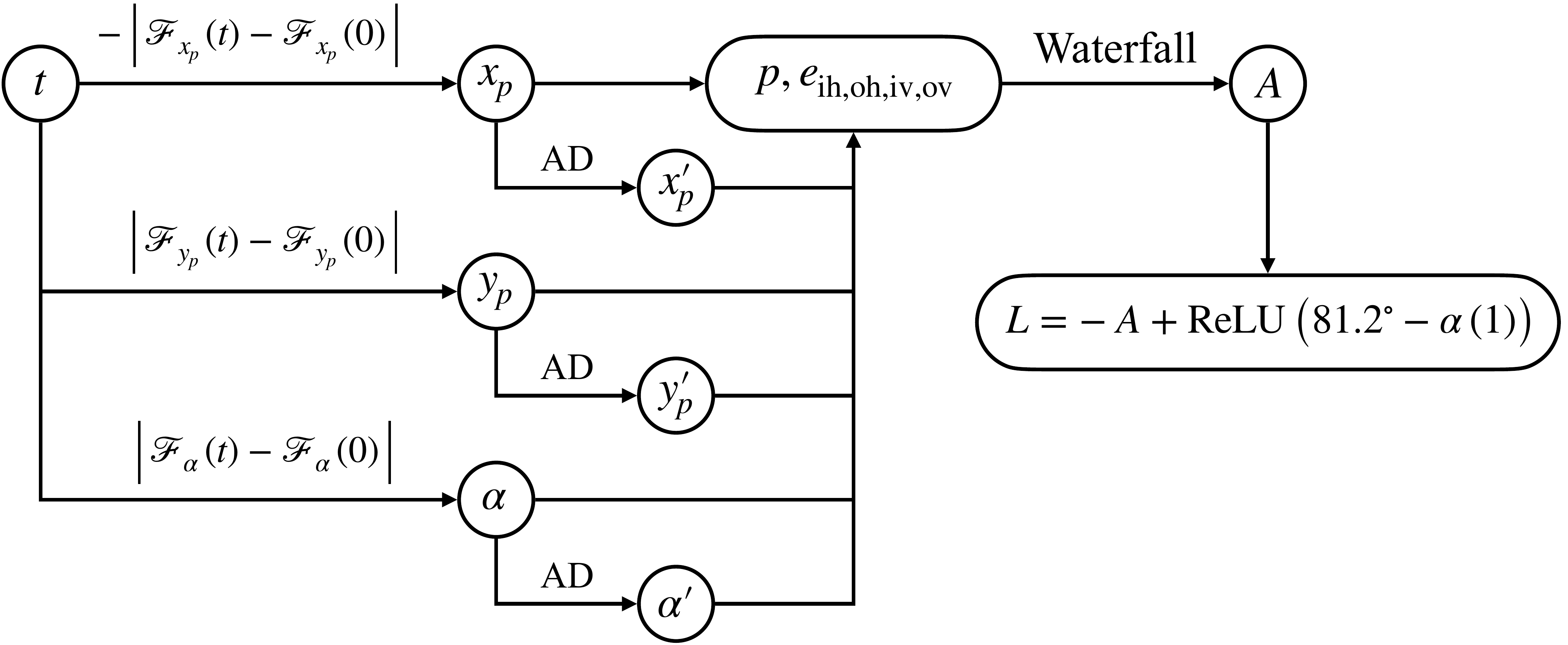}
 \caption{Our physics-informed network architecture and loss function. $\mathscr{F}$'s are unconstrained ReLU-based FCNs, and AD stands for automatic differentiation.}
 \label{fig:net}
\end{figure}

For more stable and efficient training, we make our NNs physics-informed~\cite{raissi2019physics,karniadakis2021physics}. Firstly, as indicated in Eq.~\eqref{eq:env}, the computation of the envelopes (and hence the area) depends on the derivatives of the learned functions ($x_p'$, $y_p'$ and $\alpha'$), which are efficiently calculated via automatic differentiation (AD). Secondly, we enforce the initial conditions specified in Eq.~\eqref{eq:par} by the network architecture, while the non-trivial condition in Eq.~\eqref{eq:81} is incorporated into the loss function due to its inequality nature. Besides, it is straightforward to see that a movement involving $x_p>0$, $y_p<0$, or $\alpha<0$ is forbidden by geometry. Combining all together, the functions describing a corridor movement are approximated by
\begin{equation}
 \begin{cases}
 \begin{aligned}
 x_p\left(t\right)&=-\Big|\mathscr{F}_{x_p} \left(t\right) -\mathscr{F}_{x_p} \left(0\right)\Big|,\\
 y_p\left(t\right)&=\Big|\mathscr{F}_{y_p} \left(t\right) -\mathscr{F}_{y_p} \left(0\right)\Big|,\\
 \alpha\left(t\right)&=\Big|\mathscr{F}_\alpha \left(t\right) -\mathscr{F}_\alpha \left(0\right)\Big|,
 \end{aligned}
 \end{cases}
\end{equation}
where $\mathscr{F}_{x_p}$, $\mathscr{F}_{y_p}$ and $\mathscr{F}_\alpha$ are unconstrained FCNs. With the area $A$ computed by our waterfall algorithm based on $x_p(t)$, $y_p(t)$ and $\alpha(t)$, the following loss function is adopted for backpropagation:
\begin{equation}
 L=-A+\mathrm{ReLU}\left(81.2^\circ - \alpha\left(1\right)\right),
\end{equation}
where the ReLU deactivates the penalizing term once $\alpha\left(1\right)$ reaches above $81.2^\circ$. The complete architecture is summarized in Figure~\ref{fig:net}.

\begin{figure}[t]
 \centering
 \includegraphics[width=\textwidth]{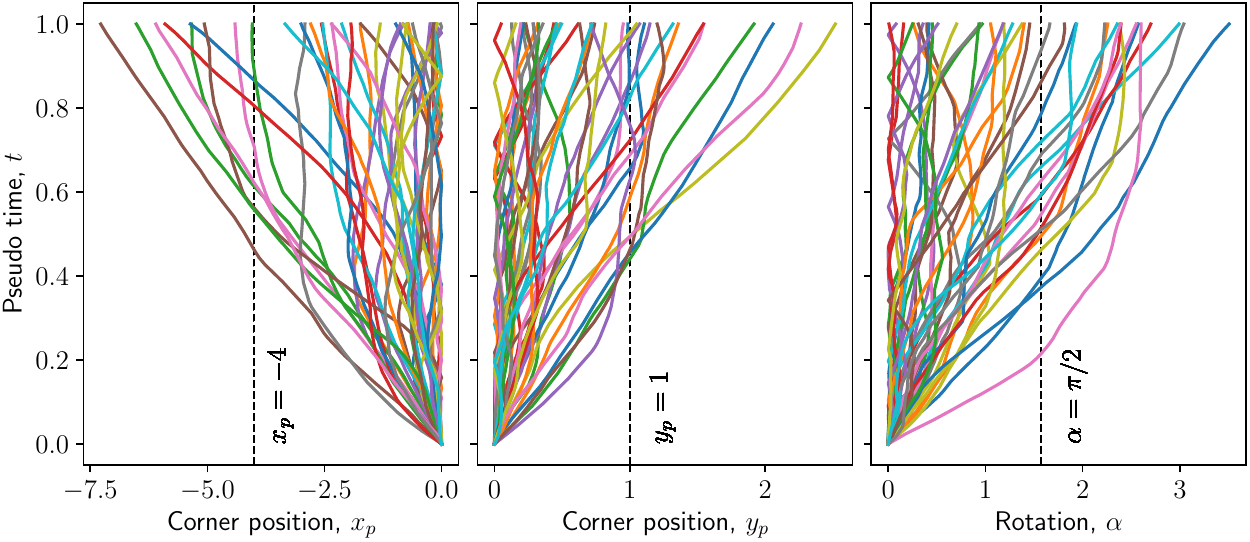}
 \caption{Function initialization with network weights sampled from scaled uniform distributions. We sample weights and biases from $\mathcal{U}\left(-s\sqrt{k}, s\sqrt{k}\right)$, where $k$ is the reciprocal of input size and $s$ a number we obtain for each function by trial and error until its admissible range is safely covered from above, as indicated by the dashed lines. Note that $s=1$ is PyTorch default.}
 \label{fig:init}
\end{figure}

NN-based optimization is gradient-based, with no guarantee on global minimum. However, the modern optimizers such as Adam~\cite{kingma2014adam,loshchilov2017decoupled}, featuring an adaptive step size and momentum, enhance the opportunity of escaping local minima. Also, the likelihood of attaining the global minimum increases if multiple approximators, initialized diversely, converge to the same local minimum. In this context, ``diverse initialization'' extends beyond mere variation in weights and biases; it implies that approximators approach their resultant minima from distinct trajectories in the parameter space. This needs meticulous scaling of the weight sampling distribution. For instance, our target function $\alpha\left(t\right)$ lies between $\left[0, \pi/2\right]$, whose approximators ($\left|\mathscr{F}_\alpha \left(t\right) -\mathscr{F}_\alpha \left(0\right)\right|$) should hence be initialized to safely cover $\left[0, \pi/2\right]$. This consideration diverges from typical machine learning practices. Figure~\ref{fig:init} illustrates a subset of our initialization samples.

\subsection{Results}
\label{sec:res}
The following training setup is adopted. We discretize the time $t\in\left[0,1\right]$ by 2000 points and distribute 10000 sources for waterfall, as determined by iterative refinement until the computed area converges to the fourth decimal place when tested on Gerver's shape. We conduct a comprehensive exploration of 6000 runs in total, encompassing 1000 random seeds for weight initialization, three initial learning rates ($10^{-3}$, $10^{-4}$ and $10^{-5}$, halved every 2000 epochs) and two network architectures (with two hidden layers of size 256 and three hidden layers of size 128). Each run undergoes 10000 epochs with Adam~\cite{kingma2014adam}. We employ double-precision floating-point format, as single precision may yield areas slightly exceeding Gerver's area near the minima, contaminating the statistics of the results. Under such configurations, each run takes about two hours on CPU; speedup by GPU is minimal due to the use of double precision. For verification, we also extend our experiments to include deeper and wider architectures, $C^\infty$ activation functions, and other weight sampling distributions. 

\begin{figure}[t]
 \centering
 \includegraphics[width=1\textwidth]{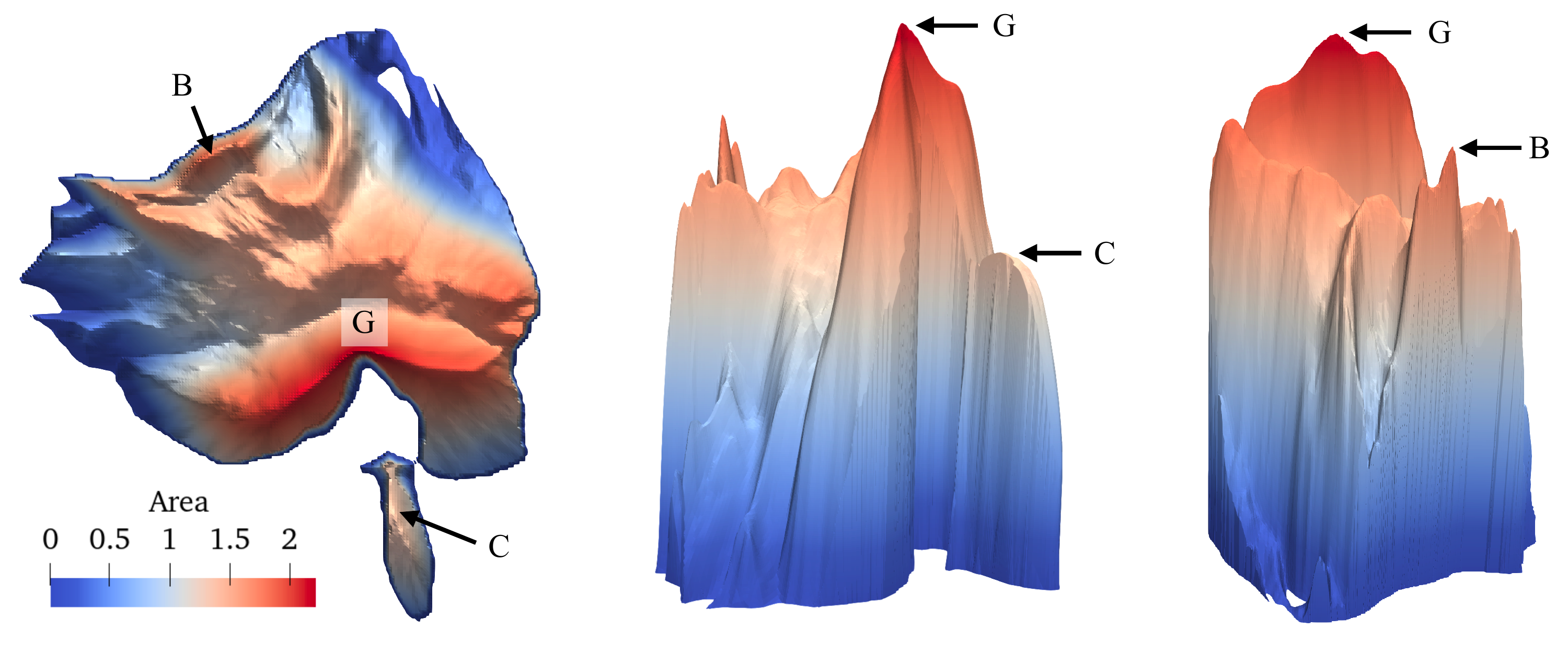}
 \caption{Landscape of sofa area. The FCNs in the visualized model have three hidden layers of size 128, trained with a learning rate of $10^{-4}$. The landscape is anchored by the dominant eigenvectors of the Hessian matrix with respect to model weights~\cite{li2018visualizing, bottcher2024visualizing}, normalized to unit length. The range of the plot is $[-1.5,1.5]\times[-1.5, 1.5]$, centered at Gerver's area (the summit). There are no other peaks in the landscape across $\left[-10,10\right]\times\left[-10,10\right]$.}
 \label{fig:aland}
\end{figure}

Among the above-defined 6000 runs, $\sim$8.5\% end up with a vanished area, mostly occurring at the largest learning rate. Excluding these outliers, the remaining trials exhibit a remarkable convergence behavior, consistently approaching a confined neighborhood surrounding Gerver's solution, which has an area of $\sim$2.219532~\cite{A128463}. The distribution of the 6000 areas most reassemble normal distribution $\mathcal{N}\left(\mu, \sigma^{2}\right)$, where $\mu = 2.219482$ and $\sigma=0.000011$, spanning a range between 2.219459 and 2.219502. Further refinement utilizing 1000 epochs of fine-tuning with the L-BFGS algorithm~\cite{liu1989limited} can improve the maximum area to 2.219515. We do not perform L-BFGS on the other models due to resource limitations. 
The models using the $\tanh$ activation function yield a strikingly similar distribution, with accelerated convergence before the area approaches the vicinity of 2.219. 
Figure~\ref{fig:aland} illustrates the Hessian-based landscape of the area for one of our converged models, offering compelling evidence for the global optimality of Gerver's area within our NN-parameterized function space.

\section{Kallus-Romik upper bound}
\label{sec:KR}

\subsection{Theory}
Yoav Kallus and Dan Romik~\cite{kallus2018improved} established an upper bound on the maximum sofa area by optimizing the rotation center at a finite sequence of rotation angles. Before presenting our findings, we first provide a brief review of their theory and discuss some visible limitations of their numerical algorithm.

Let $L_\alpha\left(\mathbf{u}\right)$ denote the set formed by rotating the corridor by angle $\alpha$ around the center $\mathbf{u}=\left(u_{1},u_{2}\right)\in\mathbb{R}^{2}$:
\begin{equation}
 \begin{aligned}
 L_\alpha\left(\mathbf{u}\right)=
 &\ \left\{(x, y) \in \mathbb{R}^2: u_1 \leq x \cos \alpha+y \sin \alpha \leq u_1+1\right\} \cap\\
 &\ \left\{(x, y) \in \mathbb{R}^2: -x \sin \alpha+y \cos \alpha \leq u_2+1\right\}\cup\\
 &\ \left\{(x, y) \in \mathbb{R}^2: x \cos \alpha+y \sin \alpha \leq u_1+1\right\}\cap\\
 &\ \left\{(x, y) \in \mathbb{R}^2: u_2 \leq-x \sin \alpha+y \cos \alpha \leq u_2+1\right\}.
 \end{aligned}
\end{equation}
Let $B\left(\beta_1, \beta_2\right)$ denote the set formed by rotating the vertical strip by angles $\beta_1$ and $\beta_2$, respectively, and taking the union of these two rotations, resulting in a butterfly-shaped region:
\begin{equation}
 \begin{aligned}
B\left(\beta_1, \beta_2\right) =
 &\ \left\{(x, y) \in \mathbb{R}^2: 0 \leq x \cos \beta_1+y \sin \beta_1\right\}\cap\\
 &\ \left\{(x, y) \in \mathbb{R}^2: x \cos \beta_2+y \sin \beta_2\leq1\right\}\cup\\
 &\ \left\{(x, y) \in \mathbb{R}^2: x \cos \beta_1+y \sin \beta_1\leq1\right\}\cap\\
 &\ \left\{(x, y) \in \mathbb{R}^2: 0\leq x \cos \beta_2+y \sin \beta_2\right\}.
 \end{aligned}
\end{equation}

\begin{figure}[t]
	\centering
	\includegraphics[width=\textwidth]{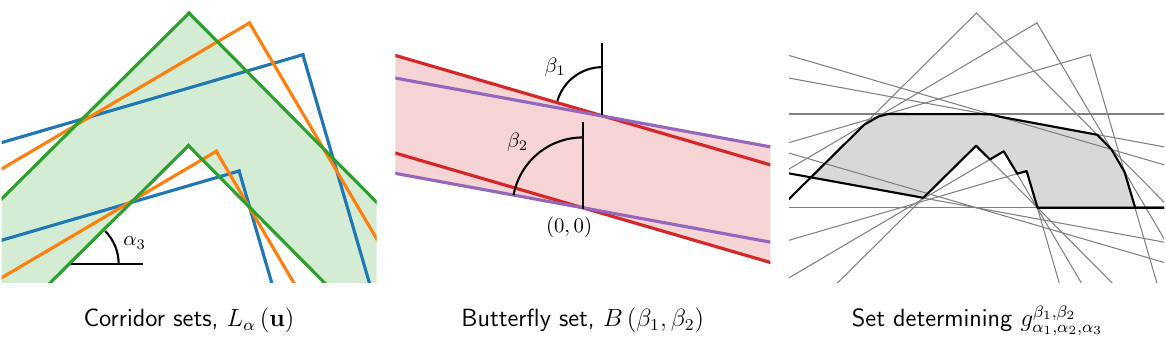}
	\caption{Sets defined for Kallus-Romik upper bound. In this example, the angles are $\alpha_1\approx16.26^\circ$, $\alpha_2\approx30.51^\circ$, $\alpha_3\approx44.76^\circ$, $\beta_1\approx73.74^\circ$ and $\beta_2\approx79.61^\circ$, corresponding to Eq.~(28) in~\cite{kallus2018improved}.}
	\label{fig:sets}
\end{figure}

Let $\lambda$ denote the area measure on $\mathbb{R}^{2}$, and $\lambda^{*}(\mathcal{X})$ the maximal area of any connected component $\mathcal{X}\subset \mathbb{R}^{2}$. Then, given a finite sequence of angles, $\left(\alpha_1, \alpha_2, \cdots, \alpha_k\right)$, and their corresponding rotation centers, $\left(\mathbf{u}_1, \mathbf{u}_2, \cdots, \mathbf{u}_k\right)$, and $\left(\beta_1, \beta_2\right)$ with $\beta_1\le\beta_2$, Kallus and Romik~\cite{kallus2018improved} first defined the following area:
\begin{equation}
g_{\alpha_1, \alpha_2, \cdots, \alpha_k}^{\beta_1, \beta_2}\left(\mathbf{u}_1, \mathbf{u}_2, \cdots, \mathbf{u}_k\right)=
	\lambda^*\left(H \cap \bigcap_{j=1}^k L_{\alpha_j}\left(\mathbf{u}_j\right) \cap B\left(\beta_1, \beta_2\right)\right),
\end{equation}
where $H =\mathbb{R}\times\left[0,1\right]$ is the horizontal strip. 
Taking the supremum of the above area with respect to the centers, they further obtained 
\begin{equation}
G_{\alpha_1, \alpha_2, \cdots, \alpha_k}^{\beta_1, \beta_2}=
\sup\left\{g_{\alpha_1, \alpha_2, \cdots, \alpha_k}^{\beta_1, \beta_2}\left(\mathbf{u}_1, \mathbf{u}_2, \cdots, \mathbf{u}_k\right): \mathbf{u}_1, \mathbf{u}_2, \cdots, \mathbf{u}_k\in\mathbb{R}^2\right\},  
\end{equation}
which they finally proved to be an upper bound of the maximum sofa area.
Note that $H\cap B\left(\beta_1, \pi/2\right)\equiv H$; namely, when $\beta_2=\pi/2$, the value of $\beta_1$ does not affect the resultant area. Therefore, the end case of $G_{\alpha_1, \alpha_2, \cdots, \alpha_k}^{\beta_1, \pi/2}$ can be simply denoted by $G_{\alpha_1, \alpha_2, \cdots, \alpha_k}$.
The above-defined sets, $L_\alpha\left(\mathbf{u}\right)$ and $B\left(\beta_1, \beta_2\right)$, and their induced $g_{\alpha_1, \alpha_2, \cdots, \alpha_k}^{\beta_1, \beta_2}$ are illustrated in Figure~\ref{fig:sets} based on a small number of angles. 

For a sofa shape that moves around the corner while rotating continuously and monotonically from 0 to some angle $\beta\in [0, \pi/2]$, define its maximum area as $A\left(\beta\right)$. The global maximum area is then $A_{\max}=\sup_{0\leq\beta\leq\pi/2}A\left(\beta\right)$. The most relevant results from Kallus and Romik~\cite{kallus2018improved} are summarized as follows:
\begin{enumerate}
 \item 
 for any $\left(\alpha_1, \alpha_2, \cdots, \alpha_k\right)$ and $\left(\beta_{1}, \beta_{2}\right)$ satisfying $\alpha_1<\alpha_2<\cdots<\alpha_k\leq\beta_1\leq\beta\leq\beta_2\leq\pi/2$, one has~\cite[Proposition 4 (iii)]{kallus2018improved}
 \begin{equation}
 A\left(\beta\right)\leq G_{\alpha_1, \alpha_2, \cdots, \alpha_k}^{\beta_1, \beta_2},
 \label{eq:prop4}
 \end{equation}
 and~\cite[Proposition 4 (ii)]{kallus2018improved}
  \begin{equation}
 A_{\max}\leq G_{\alpha_1, \alpha_2, \cdots, \alpha_k};
 \label{eq:prop4ii}
 \end{equation}

 \item given an integer $n\ge3$, let $\gamma_j=\frac{j}{n}\frac{\pi}{2}$, $j=1, 2, \cdots, n$; ~\cite[Theorem 5]{kallus2018improved} asserts that
 \begin{equation}
 A_{\max}=\lim_{n\rightarrow\infty} \ \max_{k=1,2,\cdots,\lceil n/3\rceil} \,
 G_{\gamma_1, \gamma_2, \cdots, \gamma_{n-k-1}}^{\gamma_{n-k}, \gamma_{n-k+1}},
 \label{eq:t5}
 \end{equation}
 which provides a way for constraining $A_{\max}$ asymptotically from above. 
\end{enumerate}

The above results are rigorous and elegant from a theoretical perspective. However, the numerical algorithm they present for discrete optimization of the rotation centers, $\left(\mathbf{u}_1, \mathbf{u}_2, \cdots, \mathbf{u}_k\right)$, seems to have two shortcomings: 
\begin{enumerate}
 \item their algorithm does not yield 
 $G_{\alpha_1, \alpha_2, \cdots, \alpha_k}^{\beta_1, \beta_2}$
 but another upper bound of it with unproven tightness; though this computed one is also an upper bound of $A\left(\beta\right)$, there can be space for improving the tightness if $G_{\alpha_1, \alpha_2, \cdots, \alpha_k}^{\beta_1, \beta_2}$ is evaluated directly;
 \item based on rational programming, their algorithm is computationally expensive and unscalable, which can only tackle a small number of angles. Consequently, Eq.~\eqref{eq:t5} or their Theorem 5 remains unexploited for constraining the actual value of $A_{\max}$. 
\end{enumerate}
These shortcomings motivate our work reported in the remainder of this section.

Our deep learning workflow is similar to Section~\ref{sec:learning} with two major differences. First, we use unconstrained FCNs to parameterize the functions $u_1\left(\alpha\right)$ and $u_2\left(\alpha\right)$, that is, $u_{1,2}\left(\alpha\right)=\mathscr{F}_{u_{1,2}}\left(\alpha\right)$, with no initial conditions and sign enforcement. Second, the waterfall algorithm is applied directly on the four rays in $L_\alpha\left(\mathbf{u}\right)$ and the two lines in $B\left({\beta_1,\beta_2}\right)$, instead of on envelopes. Note that envelopes do not exist because the upper bound is defined with discrete angles. These two differences indicate that our NNs are no longer informed by any ``physics'', which is consistent with the notion that the Kallus-Romik upper bound is not necessarily associated with a valid sofa movement~\cite{kallus2018improved}.

\subsection{The five-angle case}
An outstanding advantage of the Kallus-Romik upper bound is that it allows for exploring the problem at a low dimension by considering only a few angles. Kallus and Romik~\cite{kallus2018improved} defined the following five angles (we remark that they actually defined seven, but the other two are irrelevant to the sofa area)
\begin{equation}
\begin{aligned}
& \alpha_1=\arcsin \frac{7}{25} \approx 16.26^{\circ}, \\
& \alpha_2=\arcsin \frac{33}{65} \approx 30.51^{\circ}, \\
& \alpha_3=\arcsin \frac{119}{169} \approx 44.76^{\circ},\\
& \alpha_4=\arcsin \frac{56}{65}=\frac{\pi}{2}-\alpha_2 \approx 59.59^{\circ}, \\
& \alpha_5=\arcsin \frac{24}{25}=\frac{\pi}{2}-\alpha_1 \approx 73.74^{\circ}.
\end{aligned} 
\end{equation}
These angles do not have a specific geometric meaning; they are derived from $\arcsin$ functions only because the numerical algorithm of Kallus and Romik~\cite{kallus2018improved} is rational.

The inequalities below combine the optimization results from their algorithm and our NNs for $G_{\alpha_1, \alpha_2, \alpha_3, \alpha_4, \alpha_5}$ and $G_{\alpha_1, \alpha_2, \alpha_3}^{\alpha_4, \alpha_5}$, with ours in the middle:
\begin{equation}
\begin{aligned}
    G_{\alpha_1, \alpha_2, \alpha_3, \alpha_4, \alpha_5} & \leq {2.3337\pm0.0000} < 2.37, \\
    G_{\alpha_1, \alpha_2, \alpha_3}^{\alpha_4, \alpha_5} & \leq {1.9259\pm0.0000} < 2.21.
\end{aligned}
\end{equation}
Our results come from 6,000 runs, using the same hyperparameters (learning rates, architectures, seeds and epochs) as described in Section~\ref{sec:res}. The training converges so well that the variances almost vanish ($\pm0.0000$ in the above inequalities). 

\begin{figure}[b]
	\centering
	\includegraphics[width=\textwidth,trim={0 0 .75cm 1.7cm},clip]{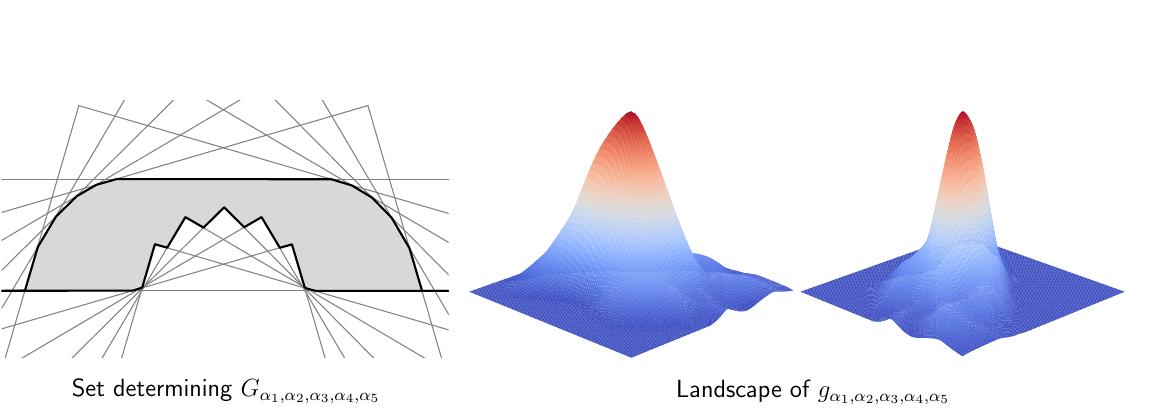}
	\caption{Set determining $G_{\alpha_1, \alpha_2, \alpha_3, \alpha_4, \alpha_5}$ ($\approx2.3337$) and landscape of $g_{\alpha_1, \alpha_2, \alpha_3, \alpha_4, \alpha_5}$. The landscape uses the rotation centers $\left(\mathbf{u}_1, \mathbf{u}_2, \cdots, \mathbf{u}_5\right)$ (instead of the network weights) as the high-dimensional variable for directional perturbation, based on the discrete nature of the Kallus-Romik upper bound (namely, the weights make less sense as the $\alpha$-sequence is prescribed). The two surface plots show the same peak from different viewpoints, with a range of $\left[-1,1\right]\times\left[-1,1\right]$ centered at $G_{\alpha_1, \alpha_2, \alpha_3, \alpha_4, \alpha_5}$ (the summit). There are no other peaks in the landscape across $\left[-10,10\right]\times\left[-10,10\right]$.}
	\label{fig:gland}
\end{figure}

Based on Eq.~\eqref{eq:prop4}, it can be shown that $A_{\max}\leq \max(G_{\alpha_1, \alpha_2, \alpha_3, \alpha_4, \alpha_5}, G_{\alpha_1, \alpha_2, \alpha_3}^{\alpha_4, \alpha_5})$. Therefore, our results suggest a tighter upper bound of 2.3337, corresponding to the set shown in  Figure~\ref{fig:gland}. The landscape of $g_{\alpha_1, \alpha_2, \alpha_3, \alpha_4, \alpha_5}$ is also shown in Figure~\ref{fig:gland}, verifying the global optimality of $G_{\alpha_1, \alpha_2, \alpha_3, \alpha_4, \alpha_5}$ in our NN-parameterized function space.

\subsection{Convergence with many angles}
The computational cost of our deep learning workflow (waterfall and backpropagation) scales with the number of angles, allowing us to directly explore the convergence theorem in Eq.~\eqref{eq:t5}. We increase the number of angles ($n$) from 10 to 10000, incremented by 10 between $\left[0, 100\right]$, by 100 between $\left(100, 3000\right]$ and by 500 afterwards, making 53 $n$'s in total. For each $n$, Eq.~\eqref{eq:t5} requires $\lceil n/3\rceil$ training instances, which are still computationally expensive for the large $n$'s. While Kallus and Romik~\cite{kallus2018improved} have shown that $81.2^\circ$ is the minimum angle the corridor must rotate for maximum area production, all current evidence hints that a rotation of $\pi/2$ should take place~\cite{gerver1992moving, romik2018differential, kallus2018improved, batsch2022numerical}, including our results reported in Section~\ref{sec:res}. Therefore, we will vary model initialization only for $k=1$ (rotation by $\pi/2$) with 20 random seeds (i.e., using one seed for $k>1$). The other hyperparameters (learning rates, architectures, and epochs) remain the same as those in Section~\ref{sec:res}. The most expensive run for $n=10000$, with 10000 sources for the waterfall and 10000 epochs, takes about ten CPU hours to complete.

\begin{figure}
 \centering
 \includegraphics[width=\textwidth]{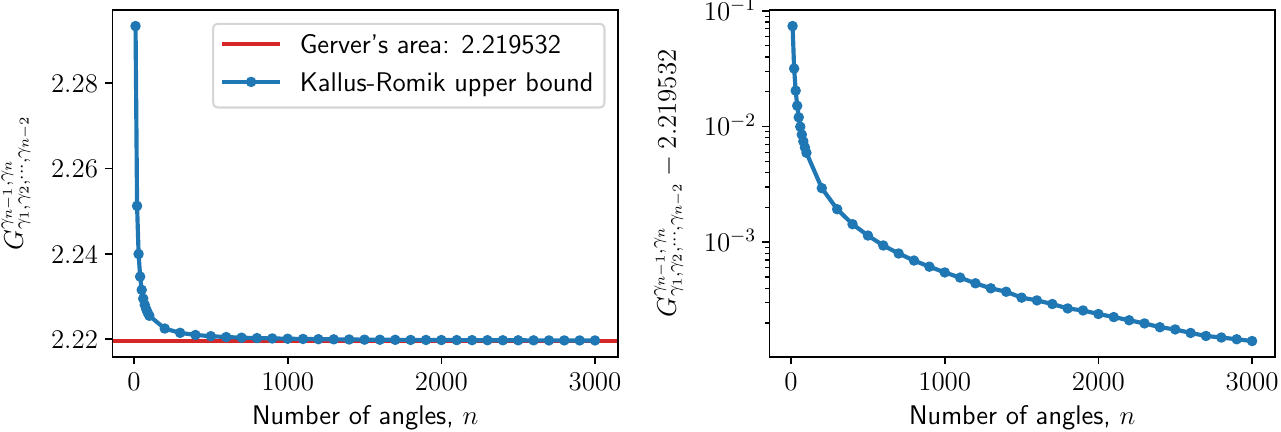}
 \caption{Convergence of $G_{\gamma_1, \gamma_2, \cdots, \gamma_{n-2}}^{\gamma_{n-1}, \gamma_{n}}$ to Gerver's area, where $\gamma_j=\frac{j}{n}\frac{\pi}{2}$, $j=1, 2, \cdots, n$. Left: $G_{\gamma_1, \gamma_2, \cdots, \gamma_{n-2}}^{\gamma_{n-1}, \gamma_{n}}$ in linear scale. Right: $G_{\gamma_1, \gamma_2, \cdots, \gamma_{n-2}}^{\gamma_{n-1}, \gamma_{n}}-2.219532$ in logarithmic scale.}
 \label{fig:conv}
\end{figure}

Our training results show that, for all the $n$'s, the maximum $G_{\gamma_1, \gamma_2, \cdots, \gamma_{n-k-1}}^{\gamma_{n-k}, \gamma_{n-k+1}}$ is achieved when $k=1$, supportive of the conjecture that the largest sofa necessitates a full rotation by $\pi/2$. We can then focus our attention on $G_{\gamma_1, \gamma_2, \cdots, \gamma_{n-2}}^{\gamma_{n-1}, \gamma_{n}}$ (i.e., $k=1$). As shown in Figure~\ref{fig:conv}, $G_{\gamma_1, \gamma_2, \cdots, \gamma_{n-2}}^{\gamma_{n-1}, \gamma_{n}}$ found by our models nicely converges to Gerver's area asymptotically from above. The relative error becomes smaller than $1\%$, $0.1\%$ and $0.01\%$ respectively at $n=30$, $300$ and $2100$, and finally reaches $0.003\%$ at $n=10000$. This result provides another piece of evidence for the global optimality of Gerver's sofa, along with the one achieved in Section~\ref{sec:res}.

\section{Discussion}
We have presented two pieces of evidence to support Gerver's conjecture that his 18-section sofa, with an area around 2.2195, is the unique shape of the maximum area that can navigate through an $L$-shaped corridor with unit width. 

Our first approach optimizes the sofa area as a function of the corridor movement parameterized by physics-informed neural networks. A critical step in our architecture is the waterfall algorithm, designed for robust, efficient, and differentiable area calculation under complex geometry of the wall envelopes. To minimize inductive bias, we emphasize the generality of our function space and the diversity of function initialization; only with such considerations can the results from extensive model training be convincing. We trained more than 6000 models; more than 90\% of them landed on Gerver's solution, with the outliers all ending up with a vanished area. 

Our second piece of evidence comes from optimizing the Kallus-Romik upper bound~\cite{kallus2018improved}. Following a similar workflow, we use neural networks to describe the rotation centers corresponding to a finite sequence of rotation angles, considering both a small and large number of angles. Using the five angles defined by Kallus and Romik~\cite{kallus2018improved}, we obtained a tighter upper bound around 2.3337, compared to their 2.37. Finally, we trained more than 6000 models based on their convergence theorem, increasing the number of angles up to 10000. The training results show that the Kallus-Romik upper bound asymptotically converges to Gerver's area, indicating that no larger sofa exists.

If a larger sofa does exist, how it has remained hidden from us for nearly 60 years? We propose two reasonable guesses in light of all previous attempts including ours. First, the shape is fractal or contains a fractal section; if this is true, its discovery would be daunting, as we are still struggling to estimate areas of known fractals. Second, the global maximum has a so small catchment basin that it can hardly be detected by optimization-based approaches (either gradient or non-gradient based). In that case, geometric insights illuminating a new movement pattern with contact mechanisms distinct from Gerver's (i.e., those prescribed in \cite{romik2018differential}) would be more essential than any algorithmic considerations.

\backmatter

\bmhead{Data availability statement}
Our code and experiments are available open-source at \url{https://github.com/kuangdai/sofa}.

\bmhead{Acknowledgments}
This work is supported by the EPSRC grant, Blueprinting for AI for Science at Exascale (BASE-II, EP/X019918/1). Computations are performed on SCARF, an HPC cluster run by Scientific Computing Department, STFC.

\bibliography{B}

\end{document}